%% file: acl2019.tex
\documentclass[11pt,a4paper]{article}
\pdfoutput=1
\usepackage[hyphens]{url}
\usepackage[hyperref]{acl2019}
\usepackage{times}
\usepackage{latexsym}
\aclfinalcopy % Uncomment this line for the final submission
\def\aclpaperid{2504}

\usepackage{amsmath,amssymb}
\usepackage{multirow}
\usepackage{color}
\usepackage{graphicx}
\usepackage{bbm}
\usepackage{xspace}
\usepackage{wasysym}
\usepackage{algorithmic}
\usepackage{float}
\usepackage[normalem]{ulem}
\usepackage{mathtools}
\usepackage{pgfplots}
\pgfplotsset{compat=1.14}
\usepackage{placeins}
\usepackage{tabularx}
\usepackage{booktabs}
\usepgfplotslibrary{groupplots}
\usepackage{xcolor}
\usepackage{colortbl}

\newcommand{\blstmavg}{\textsc{BLSTM}\xspace}
\newcommand{\wordavg}{\textsc{Word}\xspace}
\newcommand{\triavg}{\textsc{Trigram}\xspace}
\newcommand{\spavg}{\textsc{SP}\xspace}

\title{Simple and Effective Paraphrastic Similarity from Parallel Translations}

\author{John Wieting$^1$, Kevin Gimpel$^2$, Graham Neubig$^1$, and Taylor Berg-Kirkpatrick$^3$ \\
  $^1$Carnegie Mellon University,
  Pittsburgh, PA, 15213, USA \\
  $^2$Toyota Technological Institute at Chicago, Chicago, IL, 60637, USA \\
  $^3$University of California San Diego,
  San Diego, CA, 92093, USA\\
  {\small \texttt{\{jwieting,gneubig\}@cs.cmu.edu}, \texttt{kgimpel@ttic.edu}, \texttt{tberg@eng.ucsd.edu}}}

\date{}

\begin{document}
\maketitle
\begin{abstract}
We present a model and methodology for learning paraphrastic sentence embeddings directly from bitext, removing the time-consuming intermediate step of  creating  paraphrase corpora. Further, we show that the resulting model can be applied to cross-lingual tasks where it both outperforms and is orders of magnitude faster than more complex state-of-the-art baselines.\footnote{Code and data to replicate results are available at \url{https://www.cs.cmu.edu/~jwieting}.}
\end{abstract}

\section{Introduction}
\input{introduction}

\section{Learning Sentence Embeddings} \label{sec:models}
\input{models}

\section{Experiments}  \label{sec:exp}
\input{experiments}

\section{Analysis}  \label{sec:discussion}
\input{discussion}

\section{Conclusion}
\input{conclusion}

\bibliography{acl2019}
\bibliographystyle{acl_natbib}

\end{document}

%% file: introduction.tex
Measuring sentence similarity is a core task in semantics \citep{cer2017semeval}, and prior work has achieved strong results by training similarity models on datasets of paraphrase pairs~\citep{dolan-04}. However, such datasets are not produced naturally at scale and therefore must be created either through costly manual annotation or by leveraging natural annotation in specific domains, like Simple English Wikipedia~\citep{coster2011simple} or Twitter~\citep{lan-EtAl:2017:EMNLP20171-short}. 

One of the most promising approaches for inducing paraphrase datasets is via manipulation of large  bilingual corpora.
Examples include bilingual pivoting over phrases~\cite{callison2006improved,GanitkevitchDC13-short}, and automatic translation of one side of the bitext~\cite{wieting2017backtrans,wieting2017pushing,hu2019parabank}.
However, this is costly -- \citet{wieting2017pushing} report their large-scale database of sentential paraphrases required 10,000 GPU hours to generate. 

In this paper, we propose a method that trains highly performant sentence embeddings~\citep{pham-EtAl:2015:ACL-IJCNLP,hill2016learning,pagliardini2017unsupervised,mccann2017learned,conneau2017supervised} 
directly on bitext, obviating these intermediate steps and avoiding the noise and error propagation from automatic dataset preparation methods.
This approach eases data collection, since bitext occurs naturally more often than paraphrase data and, further, has the additional benefit of creating cross-lingual representations that are useful for tasks such as mining or filtering parallel data and cross-lingual retrieval. 

Most previous work for cross-lingual representations has focused on models based on encoders from neural machine translation ~\cite{espana2017empirical,schwenk2017learning,schwenk2018filtering} or deep architectures using a contrastive loss~\cite{gregoire2018extracting,guo2018effective,chidambaram2018learning}. However, the paraphrastic sentence embedding literature has observed that simple models such as pooling word embeddings generalize significantly better than complex architectures~\cite{wieting-16-full}. Here, we find a similar effect in the bilingual setting. We propose a simple model that  not only produces state-of-the-art monolingual and bilingual sentence representations, but also encode sentences hundreds of times faster -- an important factor when applying these representations for mining or filtering large amounts of bitext.
Our approach forms the simplest method to date that is able to 
achieve state-of-the-art results on multiple monolingual and cross-lingual semantic textual similarity (STS) and parallel corpora mining tasks.\footnote{In fact, we show that for monolingual similarity, we can devise random encoders that outperform some of this work.}

Lastly, since bitext is available for so many language pairs,
we analyze how the choice of language pair affects the performance of English paraphrastic representations, finding that using related languages yields the best results.

%% file: models.tex
We first describe our objective function and then describe our encoder, in addition to several baseline encoders. The methodology proposed here borrows much from past work~\cite{wieting2017pushing,guo2018effective,gregoire2018extracting,singla2018multi},
but this specific setting has not been explored and, as we show in our experiments, is surprisingly effective.

\paragraph{Training.}
The training data consists of a sequence of parallel sentence pairs $(s_i, t_i)$ in source and target languages respectively. For each sentence pair, we randomly choose a \emph{negative} target sentence $t_i'$ during training 
that is not a translation of $s_i$. Our objective is to have source and target sentences be more similar than source and negative target examples by a margin $\delta$:
\begin{align}
\min_{\theta_\textrm{src}, \theta_\textrm{tgt}} \sum_i\Big[\delta - &f_{\theta}(s_i,t_i) + f_{\theta}(s_i,t_i'))\Big]_+.
\label{eq:obj}
\end{align}
The similarity function is defined as:
\begin{align}
%\textrm{\bf Distance function:}&\\
f_\theta(s,t) = \textrm{cos}\Big(&g(s;\theta_\textrm{src}), g(t; \theta_\textrm{tgt})\Big)
\end{align} 
\noindent where $g$ is the sentence encoder with parameters for each language $\theta = (\theta_\textrm{src}, \theta_\textrm{tgt})$.
To select $t_i'$ we choose the most similar sentence in some set according to the current model parameters, i.e., the one with the highest cosine similarity.

\paragraph{Negative Sampling.} The described objective can also be applied to monolingual paraphrase data, which we explore in our experiments. The choice of negative examples differs whether we are using a monolingual parallel corpus or a bilingual parallel corpus. In the monolingual case, we select from all examples in the batch except the current pair. However, in the bilingual case, negative examples are only selected from the sentences in the batch from the opposing language.
To select difficult negative examples that aid training, we use the {\it mega-batching} procedure of \citet{wieting2017pushing}, which aggregates $M$ mini-batches to create one mega-batch and selects negative examples therefrom. Once each pair in the mega-batch has a negative example, the mega-batch is split back up into $M$ mini-batches for training.

\paragraph{Encoders.} Our primary sentence encoder simply averages the embeddings of subword units generated by \texttt{sentencepiece}~\cite{kudo2018sentencepiece}; % in the sentence; 
we refer to it as \spavg. This means that the sentence piece embeddings themselves are the only learned parameters of this model. As baselines we explore averaging character trigrams (\triavg)~\cite{wieting2016charagram} and words (\wordavg).
\spavg provides a compromise between averaging words and character trigrams, combining the more distinct semantic units of words with the coverage of character trigrams.

We also use a bidirectional long short-term memory LSTM encoder~\cite{hochreiter1997long}, with LSTM parameters fully shared between languages
, as well as \blstmavg-\spavg, which uses sentence pieces instead of words as the input tokens.  
For all encoders, when the vocabularies of the source and target languages overlap, the corresponding encoder embedding parameters are shared.
As a result, language pairs with more lexical overlap share more parameters. 

We utilize several regularization methods \citep{wieting-17-full} including dropout~\cite{srivastava2014dropout} and shuffling the words in the sentence when training \blstmavg-\spavg. Additionally, we find that annealing the mega-batch size by slowly increasing it during training improved performance by a significant margin for all models, but especially for \blstmavg-\spavg. 

%% file: experiments.tex
Our experiments are split into two groups. First, we compare training on parallel data to training on back-translated parallel data. 
We evaluate these models on the 2012-2016 SemEval Semantic Textual Similarity (STS) shared tasks~\cite{agirre2012semeval,agirre2013sem,agirre2014semeval,agirre2015semeval,agirre2016semeval}, which predict the degree to which sentences have the same meaning as measured by human judges. The evaluation metric is Pearson's $r$ with the gold labels. We use the small STS English-English dataset from \citet{cer2017semeval} for model selection.
Second, we compare our best model, \spavg, on two semantic cross-lingual tasks: the 2017 SemEval STS task~\cite{cer2017semeval} which consists of monolingual and cross-lingual datasets and the 2018 Building and Using Parallel Corpora  (BUCC) shared bitext mining task~\cite{zweigenbaum2018overview}.

\subsection{Hyperparameters and Optimization}
Unless otherwise specified, we fix the hyperparameters in our model to the following: mega-batch size to 60, margin $\delta$ to 0.4, annealing rate to 150,\footnote{Annealing rate is the number of minibatches that are processed before the megabatch size is increased by 1.} dropout to 0.3, shuffling rate for \blstmavg-\spavg to 0.3, and the size of the \texttt{sentencepiece} vocabulary to 20,000. For \wordavg and \triavg, we limited the vocabulary to
the 200,000 most frequent types
in the training data. 
We optimize our models using Adam~\cite{kingma2014adam} with a learning rate of 0.001 and trained the models for 10 epochs.

\subsection{Back-Translated Text vs. Parallel Text} \label{sec:para}

\begin{table}[t]
\setlength{\tabcolsep}{2pt}
\centering
\footnotesize
\begin{tabular} { | l | c | c | c |}
\hline
Model & \texttt{\texttt{\texttt{\texttt{\texttt{\texttt{\texttt{\texttt{en-en}}}}}}}}& \texttt{\texttt{en-cs(1M)}}  & \texttt{en-cs(2M)} \\
\hline
%\lstmavg-\spavg (20k) & 66.7 & 65.7 & 66.6 \\
%\spavg (20k) & 68.3 & \bf 68.6 & \bf 70.0 \\
%\wordavg & 66.0 & 63.8 & 65.9 \\
%\triavg & \bf 69.2 & \bf 68.6 & 69.9 \\
\blstmavg-\spavg (20k) & 66.5 & 66.4 & 66.2 \\
\spavg (20k) & 69.7 & \bf 70.0 & \bf 71.0 \\
\wordavg & 66.7 & 65.2 & 66.8 \\
\triavg & \bf 70.0 & \bf 70.0 & 70.6 \\
\hline
\end{tabular}
\caption{\label{table:backtrans} Comparison between training on 1 million examples from a backtranslated English-English corpus (\texttt{en-en}) and the original bitext corpus (\texttt{en-cs}) sampling 1 million and 2 million sentence pairs (the latter equalizes the amount of English text with the \texttt{en-en} setting). Performance is the average Pearson's $r$ over the 2012-2016 STS datasets. 
}
\end{table}

We first compare sentence encoders and sentence embedding quality between models trained on backtranslated text and those trained on bitext directly. As our bitext, we use the Czeng1.6 English-Czech parallel corpus~\cite{czeng16:2016}. We compare it to training on ParaNMT~\cite{wieting2017pushing}, a corpus of 50 million paraphrases obtained from automatically translating the Czech side of Czeng1.6 into English. We sample 1 million examples from ParaNMT and Czeng1.6 and evaluate on all 25 datasets from the STS tasks from 2012-2016. Since the models see two full English sentences for every example when training on ParaNMT, but only one when training on bitext, we also experiment with sampling twice the amount of bitext data to keep fixed the number of English training sentences.

Results in Table~\ref{table:backtrans} show two observations. First, models trained on \texttt{\texttt{\texttt{\texttt{\texttt{\texttt{en-en}}}}}}, in contrast to those trained on \texttt{en-cs}, have higher correlation for all encoders except \spavg. However, when the same number of English sentences is used, models trained on bitext have greater than or equal performance across all encoders. Second, \spavg has the best overall performance in the \texttt{en-cs} setting. It also has fewer parameters and is faster to train than \blstmavg-\spavg and \triavg. Further, it is faster at encoding new sentences at test time. 

\subsection{Monolingual and Cross-Lingual Similarity}

We evaluate on the cross-lingual STS tasks from SemEval 2017. This evaluation contains Arabic-Arabic, Arabic-English, Spanish-Spanish, Spanish-English, and Turkish-English STS datsets. These datasets were created by translating one or both pairs of an English STS pair into Arabic (\texttt{ar}), Spanish (\texttt{es}), or Turkish (\texttt{tr}).\footnote{Note that for experiments with 1M OS examples, we trained for 20 epochs.} 

\paragraph{Baselines.}
We compare to several models from prior work~\cite{guo2018effective,chidambaram2018learning}.
A fair comparison to other models is difficult due to different training setups. Therefore, we perform a variety of experiments at different scales to demonstrate that even with much less data, our method has the best performance.\footnote{We do not directly compare to recent work in learning contextualized word embeddings~\citep{peters2018deep, devlin2018bert}. While these have been very successful in many NLP tasks, they do not perform well on STS tasks without fine tuning.}
In the case of \newcite{schwenk2018filtering}, we replicate their setting in order to do a fair comparison.
\footnote{Two follow-up papers~\cite{artetxe2018margin,artetxe2018massively} use essentially the same underlying model, but we compare to \citet{schwenk2018filtering} because it was the only one of these papers where the model has been made available when this paper was written.}

As another baseline, we analyze the performance of averaging randomly initialized embeddings. We experiment with \spavg having \texttt{sentencepiece} vocabulary sizes of 20,000 and 40,000 tokens
as well as \triavg with a maximum vocabulary size of 200,000. The embeddings have 300 dimensions and are initialized from a normal distribution with mean 0 and variance 1.

\begin{table*}[th!]
\centering
\small
\begin{tabular} { | l | c | c | c | c | c | c | c | c | c |}
\hline
Model & Data & $N$ & Dim. & \texttt{\texttt{ar-ar}} & \texttt{ar-en} & \texttt{es-es} & \texttt{es-en} & \texttt{\texttt{\texttt{\texttt{\texttt{\texttt{en-en}}}}}} & \texttt{tr-en}\\
\hline
Random \triavg & OS & 1M & 300 & 67.9 & 1.8 & 77.3 & 2.8 & 73.7 & 19.4\\
Random \spavg (20k) & OS & 1M & 300 & 61.9 & 17.5 & 68.8 & 6.5 & 67.0 & 23.1\\
Random \spavg (40k) & OS & 1M & 300 & 58.3 & 16.1 & 68.2 & 10.4 & 66.6 & 22.2\\
\hline
\spavg (20k) & OS & 1M & 300 & 75.6 & 74.7 & 85.4 & 76.4 & \bf 84.5 & 77.2\\
\triavg & OS & 1M & 300 & 75.6 & \bf 75.2 & 84.1 & 73.2 & 83.5 & 74.8\\
\spavg (80k) & OS & 10M & 1024 & \bf 76.2 & 75.0 & \bf 86.2 & \bf 78.3 & \bf 84.5 & \bf 77.5 \\
\hline
\spavg (20k) & EP & 2M & 300 & - & - & 78.6 & 54.9 & 79.1 & -\\
\spavg (20k) & EP & 2M & 1024 & - & - & 81.0 & 56.4 & 80.4 & -\\
\hline
\hline
\citet{schwenk2018filtering} & EP & 18M & 1024 & - & - & 64.4 & 40.8 & 66.0 & -\\
\citet{espana2017empirical} & MIX & 32.8M & 2048 & 59 & 44 & 78 & 49 & 76 & -\\
\citet{chidambaram2018learning} & MIX & 470M/500M & 512 & - & - & 64.2 & 58.7 & - & -\\
\hline
2017 STS 1st Place & - & - & - & \bf 75.4 & \bf 74.9 & \bf 85.6 & \bf 83.0 & \bf 85.5 & \bf 77.1\\
2017 STS 2nd Place & - & - & - & \bf 75.4 & 71.3 & 85.0 & 81.3 & 85.4 & 74.2\\
2017 STS 3rd Place & - & - & - & 74.6 & 70.0 & 84.9 & 79.1 & 85.4 & 73.6\\
\hline
\end{tabular}
\caption{\label{table:multi} Comparison of our models with those in the literature and random encoder baselines. Performance is measured in Pearson's $r$ (\%). $N$ refers to the number of examples in the training data. OS stands for OpenSubtitles, EP for Europarl, and MIX for a variety of domains.
}
\end{table*}

\paragraph{Results.}
The results are shown in Table~\ref{table:multi}. We make several observations. 
The first is that the 1024 dimension \spavg model trained on 2016 OpenSubtitles Corpus\footnote{\url{http://opus.nlpl.eu/OpenSubtitles.php}}~\cite{lison2016opensubtitles2016} outperforms prior work on 4 of the 6 STS datasets. 
%Our results also show that performance increases substantially from using 1 million to using 10 million sentence pairs and is also correlated positively with dimension and size of the SP vocabulary. 
This result outperforms the baselines from the literature as well, all of which use deep architectures.\footnote{Including a 3-layer transformer trained on a constructed parallel corpus \citep{chidambaram2018learning}, a bidirectional gated recurrent unit (GRU) network trained on a collection of parallel corpora using \texttt{en-es}, \texttt{en-ar}, and \texttt{ar-es} bitext \citep{espana2017empirical},
and a 3 layer bidirectional LSTM trained on 9 languages in Europarl \citep{schwenk2018filtering}.} Our \spavg model trained on Europarl\footnote{\url{http://opus.nlpl.eu/Europarl.php}} (EP) also surpasses the model from \citet{schwenk2018filtering} which is trained on the same corpus. Since that model is based on many-to-many translation, \citet{schwenk2018filtering} trains on nine (related) languages in Europarl. We only train on the splits of interest (\texttt{en-es} for STS and \texttt{\texttt{en-de}}/\texttt{en-fr} for the BUCC tasks) in our experiments.

Secondly, we find that \spavg outperforms \triavg overall. This seems to be especially true when the languages have more \texttt{sentencepiece} tokens in common.

Lastly, we find that random encoders, especially random \triavg, perform strongly in the monolingual setting. In fact, the random encoders are competitive or outperform all three models from the literature in these cases. For cross-lingual similarity, however, random encoders lag behind because they are essentially measuring the lexical overlap in the two sentences and there is little lexical overlap in the cross-lingual setting, especially for distantly related languages like Arabic and English.

\subsection{Mining Bitext}

Lastly, we evaluate on the BUCC shared task on mining bitext. This task consists of finding the gold {\it aligned} parallel sentences given two large corpora in two distinct languages. 
Typically, only about 2.5\% of the sentences are aligned. 
Following \citet{schwenk2018filtering}, we train our models on Europarl and evaluate on the publicly available BUCC data.

Results in Table~\ref{table:bucc} on the French and German mining tasks demonstrate the proposed model outperforms \citet{schwenk2018filtering}, although the gap is substantially smaller than on the STS tasks. The reason for this is likely the domain mismatch between the STS data (image captions) and the training data (Europarl). We suspect that the deep NMT encoders of \citet{schwenk2018filtering} overfit to the domain more than the simpler \spavg model, and the BUCC task uses news data which is closer to Europarl than image captions.

\begin{table}
\centering
\footnotesize
\begin{tabular} { | l | c | c |}
\hline
Model & \texttt{en-de} & \texttt{en-fr} \\
\hline
\citet{schwenk2018filtering} & 76.1 & 74.9 \\
\spavg (20k) & 77.0 & 76.3 \\
\spavg (40k) & \bf 77.5 & \bf 76.8 \\
\hline
\end{tabular}
\caption{\label{table:bucc} F1 scores for bitext mining on BUCC.
}
\end{table}

%% file: discussion.tex
We next conduct experiments on encoding speed and analyze the effect of language choice.

\subsection{Encoding Speed} \label{sec:timing}

\begin{table}[h!]
\centering
\footnotesize
\begin{tabular} { | l | c | c |}
\hline
Model & Dim & Sentences/Sec.\\
\hline
\citet{schwenk2018filtering} & 1024 & 2,601  \\
\citet{chidambaram2018learning} & 512 & 3,049 \\
\spavg (20k) & 300 & 855,571 \\
\spavg (20k) & 1024 & 683,204 \\
\hline
\end{tabular}
\caption{\label{table:timing} A comparison of encoding times for our model compared to two models from prior work.
}
\end{table}

In addition to outperforming more complex models ~\cite{schwenk2018filtering,chidambaram2018learning}, the simple \spavg models are much faster at encoding sentences. Since implementations to encode sentences are publicly available for several baselines, we are able to test their encoding speed and compare to \spavg. To do so, we randomly select 128,000 English sentences from the English-Spanish Europarl corpus. We then encode these sentences in batches of 128 on an Nvidia Quadro GP100 GPU. The number of sentences encoded per second is shown in Table~\ref{table:timing}, showing that \spavg is hundreds of times faster. 

\subsection{Does Language Choice Matter?}

\begin{figure}
    \centering
    \includegraphics[width=0.53\textwidth]{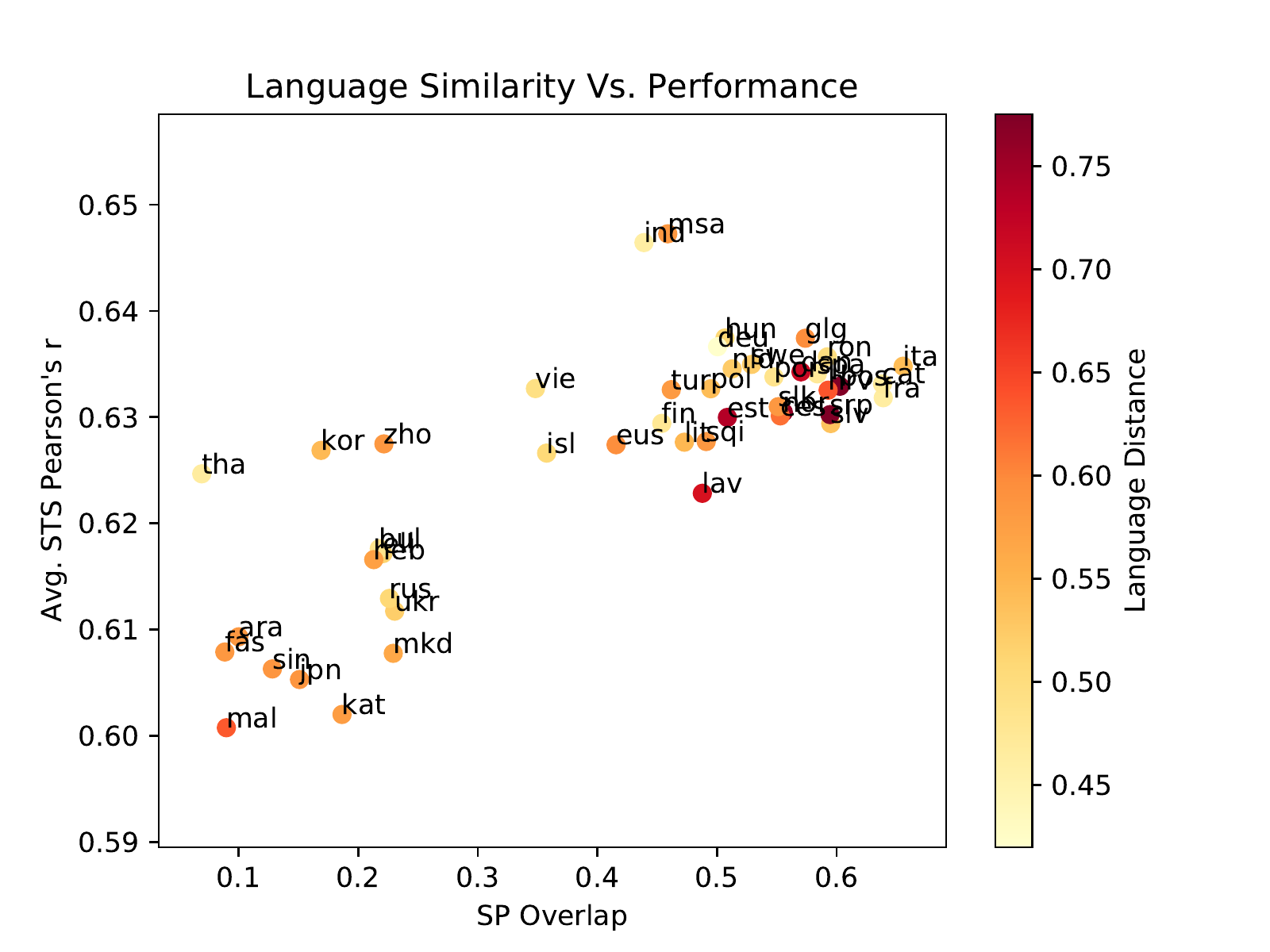}
    \caption{
    Plot of average performance on the 2012-2016 STS tasks compared to SP overlap and language distance as defined by~\citet{Littel-et-al:2017}. \label{fig:langdist}}
\end{figure}

We next investigate the impact of the non-English language in the bitext when training English paraphrastic sentence embeddings. We took all 46 languages with at least 100k parallel sentence pairs in the 2016 OpenSubtitles Corpus~\cite{lison2016opensubtitles2016} and made a plot of their average STS performance on the 2012-2016 English datasets compared to their SP overlap\footnote{We define SP overlap as the percentage of SPs in the English corpus that also appear in the non-English corpus.}
and language distance.\footnote{We used the feature distance in URIEL~\cite{Littel-et-al:2017} which accounts for a number of factors when calculating distance like phylogeny, geography, syntax, and phonology.} We segmented the languages separately and trained the models for 10 epochs using the 2017 \texttt{en-en} task for model selection. 

The plot, shown in Figure~\ref{fig:langdist}, shows that \texttt{sentencepiece} (SP) overlap is highly correlated with STS score. There are also two clusters in the plot, languages that have a similar alphabet to English and those that do not. In each cluster we find that performance is negatively correlated with language distance. Therefore, languages similar to English yield better performance. The Spearman's correlations (multiplied by 100) for all languages and these two clusters are shown in Table~\ref{table:distance}. When choosing a language to pair up with English for learning paraphrastic embeddings, ideally there will be a lot of SP overlap. However, beyond or below a certain threshold (approximately 0.3 judging by the plot), the linguistic distance to English is more predictive of performance. Of the factors in URIEL, syntactic distance was the feature most correlated with STS performance in the two clusters with correlations of -56.1 and -29.0 for the low and high overlap clusters respectively. This indicates that languages with similar syntax to English helped performance. One hypothesis to explain this relationship is that translation quality is higher for related languages, especially if the languages have the same syntax, resulting in a cleaner training signal.

\begin{table}
\centering
\footnotesize
\begin{tabular} { | l | c | c |}
\hline
Model & SP Ovl. & Lang. Distance\\
\hline
All Lang. & 71.5 & -22.8  \\
Lang. (SP Ovl. $\leq$ 0.3) & 23.6 & -63.8 \\
Lang. (SP Ovl. $>$ 0.3) & 18.5 & -34.2 \\
\hline
\end{tabular}
\caption{\label{table:distance} Spearman's $\rho \times 100$ between average performance on the 2012-2016 STS tasks compared to SP overlap (SP Ovl.) and language distance as defined by~\citet{Littel-et-al:2017}. We included correlations for all languages as well as those with low and high SP overlap with English.
}
\end{table}

We also hypothesize that having high SP overlap is correlated with improved performance because the English SP embeddings are being updated more frequently during training. To investigate the effect, we again learned segmentations separately for both languages then prefixed all tokens in the non-English text with a marker to ensure that there would be no shared parameters between the two languages. Results showed that SP overlap was still correlated (correlation of 24.9) and language distance was still negatively correlated with performance albeit significantly less so at -10.1. Of all the linguistic features, again the syntactic distance was the highest correlated at -37.5.

%% file: conclusion.tex
We have shown that using automatic dataset preparation methods such as pivoting or back-translation are not needed to create higher performing sentence embeddings. Moreover by using the bitext directly, our approach also produces strong paraphrastic cross-lingual representations as a byproduct. Our approach is much faster than comparable methods and yields stronger performance on cross-lingual and monolingual semantic similarity and cross-lingual bitext mining tasks.